\documentclass[runningheads]{llncs}
\usepackage[T1]{fontenc}
\usepackage{graphicx}
\usepackage{booktabs}
\usepackage[misc]{ifsym}

\usepackage{mwe}

\begin{document}

\title{Achieving Predictive Precision: Leveraging LSTM and Pseudo Labeling for Volvo’s Discovery Challenge at ECML-PKDD 2024}


\author{Carlo Metta\inst{1} \and
Marco Gregnanin\inst{2,3} \and
Andrea Papini\inst{4} \and Silvia Giulia Galfrè\inst{5} \and Andrea Fois\inst{6} \and Francesco Morandin\inst{6} \and Marco Fantozzi\inst{6} \and Maurizio Parton\inst{7}}

\authorrunning{Metta C. et al.}

\institute{ISTI-CNR, Italy, \email{carlo.metta@isti.cnr.it}
\and
IMT School for Advanced Studies Lucca, Italy
\and
KU Leuven, Belgium
\and
Chalmers University of Technology, Sweden
\and
University of Pisa, Italy
\and
University of Parma, Italy
\and
University of Chieti-Pescara, Italy
}

\maketitle              

\begin{abstract}
This paper presents the second-place methodology in the Volvo Discovery Challenge at ECML-PKDD 2024, where we used Long Short-Term Memory networks and pseudo-labeling to predict maintenance needs for a component of Volvo trucks. We processed the training data to mirror the test set structure and applied a base LSTM model to label the test data iteratively. This approach refined our model's predictive capabilities and culminated in a macro-average F1-score of 0.879, demonstrating robust performance in predictive maintenance. This work provides valuable insights for applying machine learning techniques effectively in industrial settings.

\end{abstract}

\section{Introduction}

The Volvo Discovery Challenge at ECML-PKDD 2024\footnote{\url{https://www.hh.se/english/about-the-university/events/discovery-challenge-ecml-pkdd-2024.html}} \cite{volvoecml}, organized in collaboration with Volvo Group Truck Technologies and Halmstad University, focused on the application of predictive maintenance for Volvo trucks. This competition tasked participants with using a large dataset from over 10,000 Volvo heavy-duty trucks to predict risk levels—Low, Medium, and High—for a specific truck component in the test set. Effective predictive maintenance is crucial for enhancing the reliability and efficiency of trucks, and it also helps reduce environmental impacts and CO2 emissions.

Our approach in this competition involved using Long Short-Term Memory (LSTM) networks, which are well-suited for time-series data like the one provided in this challenge. We combined this with pseudo-labeling to improve our model’s ability to predict on unseen data. Initially, we preprocessed the training data to closely resemble the test set. This was followed by employing a base LSTM model to assign labels to each timestep of the component’s data in the test set.

We further refined our model through an iterative process using pseudo labeling. We started with a basic LSTM model to label a subset of the test set. These labels were then used to augment our training data, creating a cycle that incrementally improved the model's accuracy with each iteration. We also implemented a boosting technique, where subsequent training phases incorporated insights from the test data labeled by earlier models.

This iterative training not only helped in enhancing the precision of our predictions but also allowed us to adapt our model effectively to the nuances of the new data. This approach is especially beneficial in fields like automotive predictive maintenance, where predicting failures accurately can significantly reduce costs and improve safety.

The methodology and results presented in this paper demonstrate a practical application of machine learning techniques in predictive maintenance. Our strategy highlights how deep learning models can be adapted for real-world industrial uses, dealing with complex and often imbalanced datasets. This paper aims to contribute to the broader discussion on how to effectively deploy machine learning models in practical scenarios, providing insights that could be useful for similar predictive maintenance tasks across different industries.

Full implementation of the exploratory data analysis, model building and training, test and evaluation is publicly available\footnote{\url{https://github.com/CuriosAI/Volvo\_Discovery\_Challenge\_ECML\_PKDD\_2024}}.

\section{Related Work}

Predictive maintenance (PdM)\cite{PdM} strategies in the automotive industry leverage machine learning to anticipate failures and optimize maintenance schedules, thereby enhancing vehicle reliability and reducing operational costs. The incorporation of advanced analytical techniques has seen a significant shift from traditional methods to data-driven approaches. In recent years, the development and application of machine learning models, such as Long Short-Term Memory (LSTM)\cite{lstm} networks and graph-based models \cite{scarselli}, have been at the forefront of this transformation.

LSTM networks, known for their effectiveness in handling time-series data, have been widely used to predict potential failures by analyzing sequential data from vehicle sensors. These models excel in capturing temporal dependencies, making them ideal for scenarios where prediction timing is crucial, as demonstrated in various industrial applications.

Graph Neural Networks (GNN)\cite{scarselli} have also been adapted to address predictive maintenance challenges, particularly in representing complex relationships within multivariate sensor data. In \cite{parton} the authors makes use of GNNs for predictive maintenance by constructing visibility graphs from time series, showcasing their potential in modeling non-linear dependencies in sensor data.

The visibility graph technique, which transforms time series into graphs to preserve their temporal characteristics, has been instrumental in this domain \cite{lacasa}. This approach facilitates the understanding of complex patterns in time series, proving particularly effective in predictive maintenance scenarios where sensor readings exhibit non-linear behaviors. Moreover, the signature method, derived from rough path theory, offers a robust framework for capturing intricate time series characteristics. It provides a rich, hierarchical summary of the series, preserving essential information across various scales \cite{signature}. This method has been applied successfully to enhance the predictive capabilities of models in high-stakes environments, such as automotive maintenance, where precise and timely predictions can prevent costly downtimes.

In addition to these advanced techniques, the integration of pseudo-labeling \cite{pseudo} and boosting strategies has emerged as a powerful approach to enhance model performance, by iteratively refining the training process using labels generated by the model itself.

Overall, the field of predictive maintenance continues to evolve, driven by advancements in machine learning and deep learning. The ongoing research and development in this area not only enhance the predictive accuracy but also significantly contribute to the sustainability and efficiency of automotive operations. As these technologies continue to mature, their integration into real-world applications promises to revolutionize the industry, leading to more reliable and cost-effective maintenance solutions.

\section{Challenge Description}

The Volvo Discovery Challenge at the ECML-PKDD 2024, hosted in collaboration with Volvo Group Truck Technologies and Halmstad University, centered around predictive maintenance with a specific focus on Volvo trucks. This annual challenge tasked participants with developing models capable of predicting the risk of failure for an undisclosed component within Volvo heavy-duty trucks. Utilizing a detailed dataset derived from real-world operational data of over 10,000 trucks, participants were expected to forecast three distinct risk levels: Low, Medium, and High, which correspond to the likelihood of component failure in the near future.

\subsection{Dataset Structure and Content}

The primary dataset provided to participants was divided into three significant files, each serving a unique purpose in the modeling process:

\begin{enumerate}
\item train$\_$gen1.csv: This file forms the backbone of the training data, containing over 157,437 readouts from 7,280 trucks. The data within this file is rich with features, spanning 308 columns. These include:
\begin{itemize}
    \item Timesteps: The time at which each readout was recorded.
    \item ChassisId$\_$encoded: An anonymized ID for each truck.
    \item gen: Indicates the generation of the component (gen1 in this dataset).
    \item risk level: The target variable indicating the risk level of component failure at each readout, categorized as Low, Medium, or High.
    \item Additional 304 feature columns: Comprising various sensor readings and operational metrics that are crucial for predicting the risk associated with the component.
\end{itemize}
The dataset exclusively contains data from the first generation of the component, tagged as gen1, and captures a snapshot at consecutive timesteps, tracking the period up until the end of data collection or component failure.
\item public$\_$X$\_$test.csv - The testing dataset mirrors the structure of the training set but introduces additional complexity:
\begin{itemize}
    \item It spans two generations of the component, gen1 and gen2, challenging the model’s ability to generalize across both familiar and new component data.
    \item Consists of 33,590 rows, each requiring a risk prediction.
    \item The data for each truck is presented in sequences fixed at 10 timesteps, adding a layer of complexity in prediction due to the truncated time window.
\end{itemize}

\item variants.csv: Contains the specifications for all trucks represented in both the training and testing datasets, encompassing:
\begin{itemize}
    \item ChassisId$\_$encoded
    \item 12 additional columns detailing encoded specifications such as engine type, cabin type, number of wheels, and number of axles. This metadata is vital for models that might correlate specific truck configurations with component reliability or failure rates.
\end{itemize}

\end{enumerate}

\subsection{Challenge Phases}

The competition was structured into two main phases. \textbf{Development Phase}: This phase allowed participants to submit up to five predictions per day. These submissions were evaluated against 20\% of the ground truth data, providing participants with frequent feedback and the opportunity to iterate on their models rapidly. \textbf{Final Phase}: This phase limited participants to three total submissions. Each submission during this phase was evaluated against the complete set of ground truth data, determining the final rankings. Only the best-performing submission from each participant was displayed on the leaderboard.

\subsection{Evaluation and Metrics}

The primary metric used to evaluate the models was the macro-average F1-score, calculated separately for predictions on gen1 and gen2 data. This metric is particularly suited for imbalanced data sets like this, as it treats all classes (Low, Medium, High) equally, irrespective of their frequency in the dataset. The final score for each participant was computed as the average of the macro F1-scores obtained for both generations, emphasizing the model's performance across both known and unseen conditions.

\subsection{Detailed Description of the Test Dataset Construction}

\begin{figure}[t]
    \centering
    \includegraphics[width=1\textwidth]{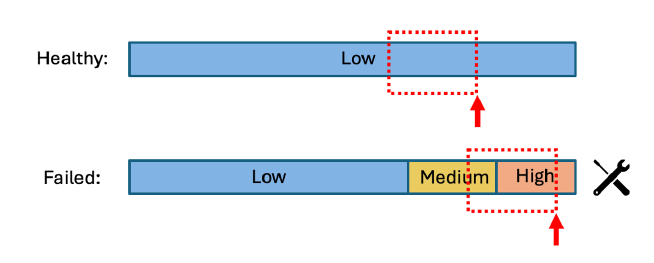}
    \caption{Sequence Extraction from Healthy and Failed components. Note that the length of all sequences is 10 time steps.}
    \label{fig:test}
\end{figure}

The public$\_$X$\_$test.csv file, crucial for model evaluation, introduces complexity not only through the inclusion of component data from two different generations (gen1 and gen2) but also in how the test sequences are specifically constructed:
\begin{itemize}
\item Each sequence in the test dataset is fixed at 10 timesteps. This uniform sequence length is designed to test the model's ability to make accurate predictions based on limited temporal data, a common scenario in real-world applications where decisions must be made quickly with constrained historical data.

\item For healthy components (those without imminent failure), a sequence of 10 timesteps is randomly selected from the available data for each ID. This random selection is intended to simulate varying operational snapshots, thereby testing the robustness of the predictive model across different operational contexts.

\item For components identified as failing (unhealthy components), the sequence extraction process is more targeted. The last timestep in the sequence is specifically chosen from a period classified as "High risk," with the preceding nine timesteps leading up to this point also included in the sequence. This ensures that the sequence captures the critical lead-up to a failure event, providing the model with data that is pivotal for identifying signs of imminent failure, see Figure \ref{fig:test}.
\end{itemize}

By limiting the sequence length, the challenge tests each model's ability to effectively utilize short-term historical data. This is particularly relevant for predictive maintenance, where early detection of potential failures can significantly enhance operational efficiency and safety, while also reducing costs associated with unscheduled maintenance and downtime.
This method of sequence generation for the test dataset plays a crucial role in the competition, as it directly influences how models are trained to generalize from the training data, which might contain longer and more varied sequences, to the test data's fixed and concise format.

By specifying these details, the challenge setup not only provides a realistic framework for the development and evaluation of predictive models but also aligns closely with typical industrial needs, where predictive maintenance must often be performed rapidly and with limited data.

\section{Methods}

\subsection{Preprocessing}

The preprocessing of the dataset was a critical initial step aimed at enhancing the quality and relevance of the data for effective model training. The first task was to clean the data by removing missing or problematic values and eliminating any empty columns which might skew the results or impair the training process.

To align the training set closer to the structure of the test set, we made several strategic decisions regarding the construction of our training data. Time series with fewer than 10 time steps were removed to ensure consistency with the test set, which exclusively includes 10-length sequences. For sequences ending with a "Low" risk label, a random 10-length subseries was extracted and added to the new dataset. Sequences ending in "Medium" were discarded to focus on more definitive outcomes (either "Low" or "High"), which provides clearer training signals for the model. For those ending in "High", a random index labeled "High" was selected to anchor a 10-length sequence ending at that index. This method not only aids in maintaining the integrity of the data by reflecting realistic operational scenarios but also addresses class imbalance by generating multiple subseries from each original sequence based on their risk level—four for every failing (High risk) truck and two for each healthy (Low risk) truck, introducing controlled randomness to balance the dataset effectively without introducing excessive correlation.
In the preprocessing phase, the data for gen1 and gen2 trucks in the test set were renormalized separately. This approach was taken to account for potential differences in data distribution between the two generations, ensuring that the model could effectively learn from and adapt to both without bias or skew.

\subsection{Model Architecture and Training}

The LSTM network was chosen due to its proficiency in handling time-series data, particularly useful for understanding sequences and their temporal dependencies. The architecture varied from 2 to 10 layers, with each layer consisting of 400 neurons, tailored to the boosting stages of training. This setup included dropout regularization at 0.5 to prevent overfitting, L2 regularization at 0.0002 to penalize large weights, and batch normalization before dropout to stabilize and speed up the training process. The model was compiled with Adam optimizer with learning rate set at 0.001 and trained using sparse categorical cross-entropy loss to directly handle multi-class labels, optimizing for accuracy.

\subsection{Semi-supervised Learning with Pseudo Labeling}

Given that the training data included only gen1 trucks while the test set comprised both gen1 and gen2, a semi-supervised learning approach with pseudo labeling was adopted. Initially, a model with 2 layers was trained on the gen1 training data and used to predict the entire test set. Post prediction, 10\% of the gen2 trucks, along with their corresponding predictions, were added to the training set, enriching it with pseudo labels. This process was iterated five times, each time with an incrementally deeper network, culminating in a 10-layer model. This strategy not only leverages the unlabeled gen2 data in the test set but also incrementally refines the model's ability to generalize across both generations.

\subsection{Post-processing Adjustments}

The post-processing stage was crucial for refining the predictions. Given that the risk in a 10-length sequence should logically increase (from Low to High) and sequences starting with "Low" should consistently predict "Low" unless proven otherwise by subsequent data points, logical rules were applied to adjust inconsistencies. For instance, sequences such as Low, Medium, ... Medium, High, ... High were adjusted to Medium, Medium, ... Medium, High, ... High.

Further adjustments were made in the distribution of predicted "High" values to mirror the expected distribution more closely. The adjustment involved slightly and randomly shifting the labels of predicted failing trucks to achieve an average distribution closer to the observed mean of 5 "High" labels, thus aligning the predictions more closely with the training set's distribution.

\subsection{Ensemble Strategy and Final Model Deployment}

To further enhance the robustness and accuracy of our predictive model, an ensemble strategy was implemented after the iterative training and pseudo-labeling process. This approach leveraged the strengths of multiple models to improve the overall performance, particularly in handling different data distribution presented by the gen1 and gen2 truck data.

\textbf{Ensemble Creation}. At the end of each 5-cycle boosting and pseudo-labeling process, we obtained five distinct models, each having evolved through five phases of incremental training—from a 2-layer to a 10-layer LSTM network. These models were not merely incremental improvements but distinct configurations trained under slightly varying data conditions, due to the inclusion of gen2 data and its associated pseudo labels.
These five final models from separate runs were then combined into an ensemble. The primary advantage of this ensemble approach lies in its ability to aggregate diverse perspectives on the data, thereby reducing the likelihood of overfitting to noise or specific artifacts in the training set and enhancing the generalization capability on the test set.

\textbf{Voting Scheme for Post-processing}. The ensemble was not only used for prediction but also played a crucial role in the post-processing adjustment phase. We employed a voting scheme where the ensemble's output was used to decide on the final adjustments necessary for the predicted sequences. This voting was particularly crucial for resolving inconsistencies and ensuring that the predictions adhered to expected logical sequences—such as ensuring non-decreasing risk levels within each 10-length sequence.

For each sequence predicted by the models in the ensemble, the majority vote was taken for each timestep to determine the most likely label. For example, sequences where a substantial majority of models predicted a uniform transition from "Medium" to "High" risk levels, but one model diverged, were adjusted according to the majority rule. This approach ensured that the final predictions were both robust and reflective of a consensus among various trained perspectives, aligning closely with the real-world dynamics modeled in the training data.

\textbf{Final Ensemble Deployment}. The deployment of this ensemble model provided a powerful tool for predictive maintenance, capable of discerning patterns and making predictions with a high degree of reliability. By combining diverse models trained through a rigorous regimen of boosting and pseudo-labeling, the ensemble could effectively handle the variability and complexity of real-world data.

\subsection{Results}

In Table \ref{fig:test} we report the final performance of the top 10 teams (out of 52 participants to the Volvo's Challenge and 791 different submission in the Development Phase) according to the whole (partially hidden) test set. For every contestant such performances are slightly lower than the one in the development phase, possibly due to a model selection that overfitted over the 20\% of the public test set. We anonymize all the other constestant team name, ID and affiliations.

\begin{table}[t]
    \centering
    \begin{tabular}{|c|c|c|c|c|}
    \hline
    \textbf{Rank} & \textbf{Team} & \textbf{Final Score} & \textbf{Score Gen1} & \textbf{Score Gen2} \\ \hline
    1 & Team1 & 0.89 & 0.89 & 0.90 \\ \hline
    \textbf{2} & \textbf{Our} & \textbf{0.88} & \textbf{0.88} & \textbf{0.88} \\ \hline
    3 & Team 3 & 0.87 & 0.87 & 0.86 \\ \hline
    4 & Team 4 & 0.78 & 0.82 & 0.74 \\ \hline
    5 & Team 5 & 0.73 & 0.78 & 0.68 \\ \hline
    6 & Team 6 & 0.68 & 0.70 & 0.66 \\ \hline
    7 & Team 7 & 0.68 & 0.69 & 0.67 \\ \hline
    8 & Team 8 & 0.68 & 0.71 & 0.65 \\ \hline
    9 & Team 9 & 0.66 & 0.68 & 0.65 \\ \hline
    10 & Team 10 & 0.66 & 0.68 & 0.64 \\ \hline
    \end{tabular}
    \caption{Top ten performances in the final phase.}
    \label{tab:result}
\end{table}

\section{Discussion and Future Work}

This paper presented a detailed exploration of LSTM networks combined with pseudo-labeling techniques in the context of predictive maintenance for Volvo trucks. The methodologies employed demonstrate the effectiveness of machine learning techniques in industrial settings, particularly in automating and improving maintenance strategies for heavy-duty trucks. Our approach, which culminated in a significant macro-average F1-score, reflects the robust performance of our models and emphasizes the potential for further applications of deep learning in predictive maintenance.

As we look toward future industrial applications, our intention is to integrate several advanced techniques and innovations that have shown promise in recent research. For instance, exploring methods to increase biases rather than weights in neural networks, as discussed in \cite{dac}, offers a novel perspective on enhancing model efficiency and generalization capabilities without the cost of significantly increased computational resources. Similarly, approaches like those described in \cite{switchpath}, which detail the SwitchPath method to enhance exploration in neural networks, are of particular interest, although the compatibility of such methods with LSTM architectures and their typical activation functions like tanh, rather than ReLU, remains an area for further investigation and experimentation.

Moreover, the potential for incorporating architectures such as transformers and stochastic transformers is compelling. These models have revolutionized fields like natural language processing and are beginning to show promise in other domains, including vision and sequential data processing. The ability of transformers to handle long-range dependencies may offer a significant advantage in predictive maintenance, where patterns over time are critical to accurate predictions. Our ongoing work includes experimenting with these new architectures, focusing on how they can be adapted and integrated into existing frameworks used for predictive maintenance.

In conclusion, while this study has provided substantial insights and advanced the application of machine learning in predictive maintenance, there are plenty of opportunities for further exploration. Future work will aim to refine these approaches, testing their limits and capabilities in industrial applications to ensure they can meet the demands of real-world environments. The integration of innovative techniques will continue to drive forward the capabilities of predictive maintenance systems, reducing downtime and maintenance costs while enhancing reliability and efficiency.

\begin{credits}
\subsubsection{\ackname}

This work is partially funded by EU Horizon 2020: G.A. 871042 SoBig-Data++, NextGenEU - PNRR-PEAI (M4C2, investment 1.3) FAIR and “SoBigData.it”, and by GNSAGA INdAM group.

\subsubsection{\discintname}
The authors have no competing interests to declare that are
relevant to the content of this article.
\end{credits}
%
%
%
\bibliographystyle{splncs04}
\bibliography{references}
%
\end{document}